\documentclass[runningheads]{llncs}
\usepackage[T1]{fontenc}
\usepackage{graphicx}
\usepackage[utf8]{inputenc}
\usepackage[english]{babel}
\usepackage[caption=false]{subfig}
\usepackage{amsmath}
\usepackage{amsfonts}
\usepackage{amssymb}
\usepackage{graphicx}
\usepackage{algorithmicx}
\usepackage{algcompatible}
\usepackage{algorithm2e}
\usepackage{placeins} 
\begin{document}
\title{Exploring and Learning Structure: \\Active Inference Approach in Navigational Agents}
\titlerunning{Exploring and Learning Structure}
%
\author{Daria de Tinguy\inst{1}\orcidID{0000-0003-1112-049X} \and
Tim Verbelen\inst{2} \and
bart Dhoedt\inst{1}}
\authorrunning{D. de Tinguy \& al}
%
\institute{Ghent University, Ghent, Belgium
\email{first\_name.family\_name@ugent.be}\\
\and
Verses AI 
\email{tim.verbelen@verses.ai}}
\maketitle              
\begin{abstract}
Drawing inspiration from animal navigation strategies, we introduce a novel computational model for navigation and mapping, rooted in biologically inspired principles. Animals exhibit remarkable navigation abilities by efficiently using memory, imagination, and strategic decision-making to navigate complex and aliased environments. Building on these insights, we integrate traditional cognitive mapping approaches with an Active Inference Framework (AIF) to learn an environment structure in a few steps. Through the incorporation of topological mapping for long-term memory and AIF for navigation planning and structure learning, our model can dynamically apprehend environmental structures and expand its internal map with predicted beliefs during exploration. Comparative experiments with the Clone-Structured Graph (CSCG) model highlight our model's ability to rapidly learn environmental structures in a single episode, with minimal navigation overlap. this is achieved without prior knowledge of the dimensions of the environment or the type of observations, showcasing its robustness and effectiveness in navigating ambiguous environments.

\end{abstract}
\section{Introduction}

A functional navigation system must seamlessly fulfil three key functions: self-localisation, mapping, and path planning. This requires both a sensing component for spatial perception and a storage capability to extend these perceptions temporally and spatially~\cite{Human_rodent_spatial_rep}. Animals exhibit a remarkable capacity for rapidly learning the structure of their environment, often in just one or a few visits, relying on memory, imagination, and strategic decision-making~\cite{few_one_shot_learning,mice_in_labyrith}.

The hippocampus and neocortex play crucial roles in episodic memory, spatial representation, and relational inference. Mammals rely on mental representations of spatial structures, traditionally viewed as either cognitive maps or cognitive graphs, conceptualising environmental space as a network of nodes~\cite{Human_rodent_spatial_rep,cscg_space_latent_seq,humans-cognitive-map,humans-hierarchic-plan-subway}. Recent research suggests an integrated approach combining these concepts is more effective~\cite{map-graph-cognition}.

Our approach adopts this viewpoint, proposing a topological map incorporating internal motion (Euclidean parameters) to delineate spatial experiences. The neural positioning system, found in rodents and primates, supports self-localisation and provides a metric for distance and direction between locations~\cite{Human_rodent_spatial_rep}. This system includes place cells, heading direction cells~\cite{heading_cells_humans}, grid cells~\cite{grid_cell_nav}, speed cells~\cite{speed_cells}, and border cells~\cite{border_cells}, working together to enable rapid learning, disambiguation of aliases, and a comprehensive understanding of spatial navigation~\cite{grid_place_cells_clustered_envs}.

\begin{figure}[!htb]
    \centering
    \includegraphics[width=9.2cm]{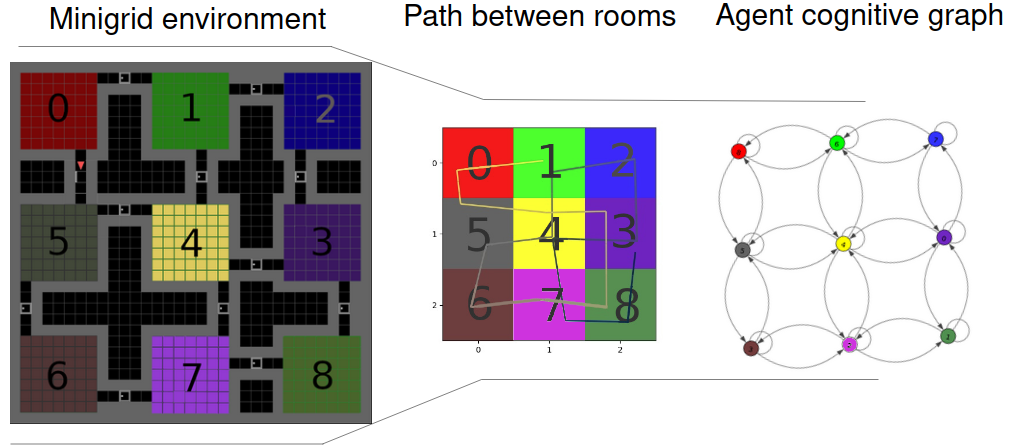}
    \caption{From the mini-grid environment~\cite{gym_minigrid,ours_2024} with different rooms annotated by colour to the path our agent took -from black to white- to form a successful exploration correctly linking all the rooms up to the agent's internal topological graph with the state associated to each room.}
    \label{img:from_env_to_topo}
\end{figure}

Building on these concepts, we introduce a novel model that dynamically learns environmental structure and expands its cognitive map. Integrating visual information and proprioception (inferred body motion), our model constructs locations and connections within its cognitive map. Starting with uncertainty, the model envisions action outcomes, expanding its map by incorporating hypotheses into its generative model, analogous to Bayesian model reduction~\cite{bayesian_model_reduc} that grows its model upon receiving new observation, we extend ours upon predicted beliefs. This process allows our model to efficiently navigate and comprehend environmental structures with minimal steps, using an active inference navigation scheme~\cite{nav_aif}. Compared to the Clone-Structured Graph (CSCG)~\cite{cscg_structuring_knowledge}, our model rapidly learns environmental layouts more efficiently. An exploration run is shown in Figure~\ref{img:from_env_to_topo}, displaying from left to right the full environment, the extracted observation and exploration path and the resulting internal map of the agent.

\section{Related work}

While navigating their environment, animals often encounter ambiguous sensory inputs due to aliasing, resulting in repetitive observations, such as encountering two overly similar corridors. They must rapidly disambiguate the structure of their environment to navigate successfully.

Models like the Clone-Structured Graph (CSCG)~\cite{cscg_abstraction} or Transformers representations~\cite{cscg_transformers} have been proposed to form cognitive maps that disambiguate aliased environments through partial observations. However, these models require substantial training time using random or hard-coded policies. In contrast, animals adapt their actions based on subtle cues and incentives, learning to navigate with minimal instances~\cite{few_one_shot_learning}.

Animals exhibit decision-making abilities powered by imagination (estimating actions' consequences) and a holistic understanding of the environment, naturally imagining un-visited areas and guiding their next steps~\cite{grid_cell_nav}. This intuitive decision-making process, considering incentives like food or safety, rapidly directs them toward their objectives~~\cite{mice_in_labyrith}. 

Integrating observations with proprioception~\cite{Human_rodent_spatial_rep} helps animals circumvent aliasing, using a process similar to active inference for judgement~\cite{nav_aif}. Active inference involves continuously updating internal models based on sensory inputs, enabling adaptive and efficient decision-making. This normative framework explains cognitive processing and brain dynamics by positing that actions and perceptions aim to minimise free energy, encapsulating causal relationships among observable outcomes, actions, and hidden states~\cite{AIF_book,AIF_learning}.

At the core of adaptive behaviour is the balance between exploitation (selecting the most valuable option based on existing beliefs) and exploration (choosing options that facilitate learning)~\cite{curiosity_exploitative}. Recent behavioural evidence suggests humans mix random and goal-directed exploration~\cite{human_exploration}. Our model adopts this balance through free energy minimisation, choosing stochastic policies to enhance environmental understanding. This enables active learning, rapidly reducing uncertainty over model parameters (i.e. reducing uncertainty over our beliefs)~\cite{curiosity_exploitative}.

By balancing curiosity and goal-directed behaviour through free energy, our system guides an agent meaningfully and learns in a biologically plausible way~\cite{AIF_book}. It achieves few-shot or one-shot learning, similar to mice in a labyrinth~\cite{mice_in_labyrith}. By projecting the consequences of actions into its internal map, the agent extends its imagination beyond known territories, improving its navigation ability to explore environments of any dimension.

\section{Method}

In our study, our agent initiates exploration of the environment without any prior knowledge regarding the observations and dimensions of the map it is about to navigate. 
Subsequently, we will clarify how, at each step, the agent engages in inferring the current state, a process that integrates both the notion of observation and proprioception (position perception given a motion). This inference task involves updating past beliefs based on the latest observation and motion, following the principles of a Partially Observable Markov Model (POMDP). Henceforth, the agent strategically envisions sequences of actions to explore, termed policies, while concurrently expanding its internal map to accommodate potential unexplored areas with uncertain priors. Although the agent may know the relative positions of these areas, it does not foresee observations. This iterative and multi-step process serves as the cornerstone for the agent's adaptive learning and navigation strategies within the environment. 

\subsection{Inference and spatial abstraction}

\begin{figure}[!t]
    \centering
    \includegraphics[width=4.5cm]{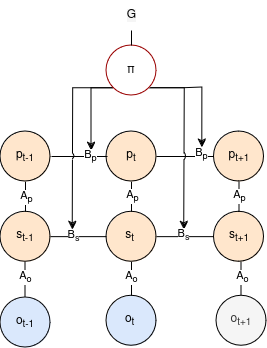}
    \caption{factor graph POMDP of our generative model transitioning from past and present (up to time-step $t$) to future (time-step $t+1$). The pose $p_t$ is inferred from the previous pose $p_{t-1}$ and the action from policy $\pi$, while the state $s_t$ determined by the corresponding observation $o_t$ and influenced by the previous state $s_{t-1}$, pose $p_{t}$ and action $a_{t-1}$. Past actions and observations are assumed observable, indicated by a blue colour. In the future, the actions are defined by a policy $\pi$ influencing the new states and position in orange and new predictions in grey.}
    \label{img:pomdp}
\end{figure}

In the context of Active Inference (AIF), the process of inferring the agent's current state involves integrating sensory inputs and prior beliefs within a Partially Observable Markov Decision Process (POMDP). We consider that our inference mechanism operates at the highest level of abstraction within a hierarchical spatial framework~\cite{ours_2024}, where lower layers handle observation transformation and the concept of blocked paths, akin to how visual observations are processed in the visual cortex and motion limitations are perceived by border cells~\cite{border_cells}. Figure~\ref{img:from_env_to_topo} illustrates the agent navigating through an environment, where doors signify transitions to different states while walls correspond to obstacles. We give our agent the notion that doors lead to another location, while walls lead to the same observation and the pose stays static. At the centre of this Figure, we see a path taken by the agent depicted along with the observations perceived by our model, demonstrating how observations are simplified and generalised at the highest abstraction level into a single colour per room (floor colour). The internal topological map generated by the agent based on its exploration path is presented in the final frame of  Figure~\ref{img:from_env_to_topo}.
The underlying POMDP model guiding this inference process is depicted in Figure~\ref{img:pomdp}, where the current state $s_t$ (defining a room) and position $p_t$ (the location of that room) are inferred based on the previous state $s_{t-1}$, $p_{t-1}$ and action $a_{t-1}$ leading to the current observation $o_t$ (the colour of that room). The generative model capturing this process is described by Equation~\ref{eq1}, where the joint probability distribution over time sequences of states, observations, and actions is formulated. Tildes are used to denote sequences over time.
\begin{equation} 
P(\tilde{o}, \tilde{s}, \tilde{p} ,\tilde{a}) = P(o_0| s_{0})P(s_0)P(p_0)P(a_0) \prod_{t=1}^\tau 
P(o_t| s_{t})P(s_t, p_t|s_{t-1},p_{t-1},a_{t-1})  
\label{eq1}
\end{equation} 
Due to the posterior distribution over a state becoming intractable in large state spaces, we use variational inference instead. This approach introduces an approximate posterior denoted as $Q(\tilde{s},\tilde{p}| \tilde{o}, \tilde{a})$ and is presented in equation~\ref{eq2}~\cite{aif_step_by_step}. 

\begin{equation} 
Q(\tilde{s},\tilde{p}| \tilde{o}, \tilde{a}) = Q(s_0, p_0| o_0) \prod_{t=1}^\tau
Q(s_t, p_t| s_{t-1},p_{t-1}, a_{t-1}, o_t )  
\label{eq2}
\end{equation} 


The classical inference scheme heavily relies on past and current experiences to localise the agent within its environment, using observation alone, the agent would be weak to aliased observations at different locations. By combining observation with the agent's proprioception the model is much more robust in differentiating ambiguous environments.  The internal positioning $p_0$ is initialised at the start of exploration in the absence of prior information, and is updated as the agent transitions between rooms (e.i., by passing through a door), thus as long as the agent is confident in its current state. the POMDP factor graph showing the association between poses, states and observations is illustrated in Figure~\ref{img:pomdp}

If the agent were to be kidnapped and re-localised elsewhere, the observation $o$ and inferred position $p$ would not match expectations and the confidence in the state would decrease. If the confidence in the state goes below a given threshold, the agent stops updating its internal model given new information and focuses on re-gaining confidence over its state/location. 

However, inferring the position $p$ has much more to offer than localisation robustness, it is key to extending the internal map over unexplored areas yet to be integrated into the model through parameter learning. 

\subsection{Parameter Learning}

Learning within the Active Inference framework encompasses the adaptation of beliefs concerning model parameters, such as transition probabilities $P(s_t|s_{t-1})$ (e.g. how rooms are connected) and likelihood probabilities $P(o_t|s_t)$ (e.g. what a room looks like). These parameters reflect the structural connectivity of the environment and the expected sensory outcomes given particular states.

Generative models in Active Inference rely on prior beliefs regarding parameter distributions, with updates driven by the active inference framework~\cite{AIF_book}. Unlike traditional discrete-time POMDP, where either transitions or likelihoods probabilities are fixed and updating parameters implies reasoning over a fixed spatial dimension~\cite{weird_HAIF,aif_scene_construction,Toon_cognitive_maps}, our model learns the probabilities of all its Markov matrices and extends their dimensions dynamically.   
The state transitions  $B_s =P(s_t|s_{t-1},a_{t-1})$ and the observation $A_o=P(o_t|s_t)$, position likelihood$A_p=P(p_t|s_t)$ probabilities are optimised over transitions.
The position transition $B_p=P(p_t|p_{t-1},a_{t-1})$, however, is not a Markov matrix and entails an incremental process based on consecutive motions (experimented or predicted), without any parameters to be learned by belief optimisation. 

The optimisation of beliefs of the generative model parameters $\theta$ occur after state inference and involves minimising the free energy $F_\theta$ while considering prior beliefs and uncertainties associated with both parameters and policies, as defined in~\cite{AIF_book}:
\begin{equation}
\begin{aligned}
    \theta =& (A_o,A_p,B_s)\\
    F_\theta =& \mathbb{E}_{Q(\pi,\theta)}[F(\pi,\theta)] + D_{KL}[Q(\theta)||P(\theta)] + D_{KL}[Q(\pi)||P(\pi)]
\end{aligned}
\end{equation}


With $P$ and $Q$ being respectively the joint distribution and the approximate posterior of the model.  The model updates its parameters based on observed data and transitions, expanding the observation dimension of A upon encountering new information as can be seen in \cite{AIF_modeling_struct}. At initialisation, high certainty is assigned to the likelihood probabilities integrating the first observation. After realising the parameter update based on priors, the model edits its internal map dimensions and parameters based on predicted transitions, expanding all parameters in their state dimensions, thus improving exploration in unexplored environments of any unknown size. 



\subsection{Incorporating spatial dynamics in model parameters}

To extend the internal map (our state space), we propose a novel approach where the agent predicts one-step policy outcomes in all directions considering detected obstacles. $B_p$ can expand its position dimension given a motion and $A_p$ considers the probability of being at a given state given the position. When we predict a new position given no obstacle in a direction, $B_p$ expands.
If the expected motion leads to an un-visited location, $s_{t+1}$ does not exist in the model. The state is undefined while the position $p_{t+1}$ is certain, therefore all Markov matrices are expected to grow in their state dimension to match this new prediction.
This process enables the dynamic expansion of the dimensions of both the observation and position likelihoods ($A_o$, $A_p$) and state transition ($B_s$) to consider the novel state $s_{t+1}$ in a process equivalent to~\cite{bayesian_model_reduc}. Subsequently, the state transition probability ($B_s$) and position likelihood ($A_p$) can be updated through the same equation~\ref{eq:B_up}, here shown with a transition matrix.
\begin{equation}
    B_\pi = Q(s_{t+1}|s_{t}, \pi) Q(s_{t}) * B_\pi  * learning\_rate
    \label{eq:B_up}
\end{equation}
With $s_{t+1}$ being a new state if there are no obstacles detected and a new position is predicted or the same state $s_t$ otherwise. The learning rate is set higher for experimented transitions than imagined transitions such that we form new connections weaker toward expected places compared to visited places as we would expect from animal synaptic learning~\cite{hippo_cognitive_map}. $A_o$ has grown in its state dimension, however, it lacks information regarding the specific observation $o_{t+1}$ expected in that location, resulting in a uniform distribution for that state. Such areas exhibit high uncertainty in their observation likelihood model. Using those prior, the agent can leverage the Active Inference scheme to determine where to direct itself to maximise its objective (e.g. forming a comprehensive map of the environment). 
While previous models such as~~\cite {bayesian_model_reduc,AIF_modeling_struct} adjust their internal model growth to accommodate new patterns of observations, we extend the concept to predict, un-visited, areas and generate new states holding no observation. Those unknown states are therefore highly attractive when seeking information gain and largely improve exploration strategy. 




\subsection{Policy Selection in Active Inference}


Policy selection plays a crucial role in exploring those expected states generated by the model. The AIF guides the agent's decision-making process based on the minimisation of expected surprise and uncertainty. Policy selection, informed by the AIF, determines the agent's actions and map extension in response to sensory inputs and internal beliefs.

Typically, agents are assumed to desire to minimise their variational free energy ($F$), which can serve as a metric to quantify the discrepancy between the joint distribution  $P$ and the approximate posterior $Q$ as presented in Equation~\ref{eq:F}.

\begin{equation}
\begin{aligned}
F & = \mathbb{E}_{Q(\tilde{s},\tilde{p}|\tilde{a},\tilde{o})}[\log[Q(\tilde{s},\tilde{p}|\tilde{a},\tilde{o})] - \log[P( \tilde{s},\tilde{p},\tilde{a},\tilde{o})] \\
 & = \underbrace{D_{KL}[Q(\tilde{s},\tilde{p}|\tilde{a},\tilde{o}))||P(\tilde{s},\tilde{p}|\tilde{a},\tilde{o})]}_\text{posterior approximation} -\underbrace{\log[ P(\tilde{o})]}_\text{log evidence}\\
 & = \underbrace{D_{KL}[Q(\tilde{s},\tilde{p}|\tilde{a},\tilde{o}))||P(\tilde{s},\tilde{p},\tilde{a})]}_\text{complexity} - \underbrace{\mathbb{E}_{Q(\tilde{s},\tilde{p}|\tilde{a},\tilde{o})}[\log[P(\tilde{o}| \tilde{s})]}_\text{accuracy}
\end{aligned}
\label{eq:F}
\end{equation}

Active inference agents aim to minimise their free energy by engaging in three main processes: learning, perception, and planning. Learning involves optimising the model parameters, perception entails estimating the most likely state, and planning involves selecting the policy or action sequence that leads to the lowest expected free energy. Essentially, this means that the process involves forming beliefs about hidden states that offer a precise and concise explanation of observed outcomes while minimising complexity.

While planning, however, we use the expected free energy (G), indicating the agent's anticipated variational free energy following the implementation of a policy $\pi$. Unlike the variational free energy, which focuses on current and past observations, the expected free energy incorporates future expected observations generated by the selected policy.

\begin{equation}
\begin{aligned}
    G(\pi,\tau) =& \underbrace{\mathbb{E}_{Q(o_\tau,s_\tau|\pi)}[log(Q(s_\tau|\pi) - log(Q(s_\tau| o_\tau, \pi))]}_\text{information gain term} \\ 
    &- \underbrace{\mathbb{E}_{Q(o_\tau,s_\tau|\pi))}[log(P(o_{\tau}))]}_\text{utility term}
\end{aligned}
\label{eq:planning_as_inf}
\end{equation}
The expected information gain quantifies the anticipated shift in the agent's belief over the state from the prior $Q(s_\tau|\pi)$ to the posterior $Q(s_\tau| o_\tau,\pi)$ when pursuing a particular policy.
On the other hand, the utility term assesses the expected log probability of observing the preferred outcome under the chosen policy. This value intuitively measures the likelihood that the policy will guide the agent toward its prior preferences. In this study, we give no prior preference to the agent, as it does not know the environment (unknown observations and map size).

To calculate this expected free energy $G(\pi)$ over each step $\tau$ of a policy we sum the expected free energy of each time-step. 
\begin{equation}
    G(\pi) = \sum_\tau G(\pi,\tau)
\end{equation}

To consider the best policy, we recall that active inference achieves goal-directed behaviour by selecting policies minimising this expected free energy, thereby aiming to produce observations closer to preferred outcomes or prior preferences. This is achieved by setting the approximate posterior over policies as in Equation~\ref{eq:P(pi)}~\cite{Toon_cognitive_maps}:
\begin{equation}
    P(\pi) = \sigma (-\gamma G(\pi))
\label{eq:P(pi)}
\end{equation}
Where $\sigma$, the softmax function is tempered with a temperature parameter $\gamma$, given as a hyper-parameter, converting the expected free energy of policies into a categorical distribution over policies. Actions are then sampled based on this posterior distribution, with lower temperatures resulting in more deterministic behaviour.

By navigating without a clear preference, we desire the highest information gain, effectively pushing the agent toward states it anticipates but doesn't know what to expect from. 

\section{Results}

We explore experimental scenarios where an agent navigates within a grid environment with cardinal motions and still motion. The agents have no direct access to a map of the environment and visual observations are considered to undergo hierarchical processing, transforming them from a vector to a single descriptor corresponding to one colour per room.  They receive localised sensory inputs, corresponding to the current room they are in. Sensory inputs which are possibly repeated at different locations (aliased observations). Given a series of discretised egocentric observations and actions, the agent must deduce the latent topology of its environment to assess various navigation options. Learning this latent graph from aliased observations presents a challenge for most artificial agents~\cite{SLAM_alias}.
We contrast our model with CSCG~\cite{cscg_pres}, a specialised variant of Hidden Markov Models (HMM). CSCG employs a probabilistic approach, using sequences of action-observation pairs without assuming Euclidean geometry. Each observation corresponds to a subset of hidden states known as clones. Although these states share the same observation likelihood, they differ in their implied dynamics encoded in the transition model. By analysing the sequence of action-observation pairs, specific clones with higher likelihoods can disambiguate the aliased observations. Initially, CSCG gathers a dataset through maze exploration to learn the spatial structure~\cite{cscg_pres}.  

\begin{figure*}[!htb]

\hfill 
   \subfloat[average steps to explore the environments]{\label{img:average_steps_models_per_env}
      \includegraphics[width=.9\textwidth]{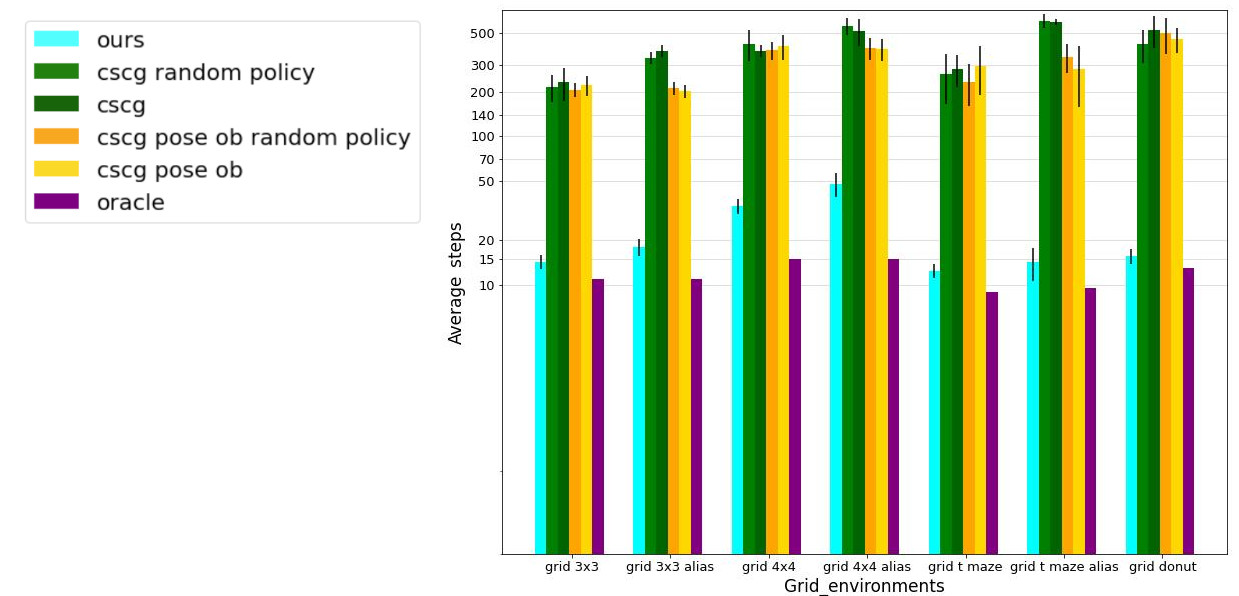}}
\hspace{1em} 

\hfill 
   \subfloat[average steps to discover the environments]{\label{img:average_steps_models_room_discovery_per_env}
      \includegraphics[width=.6\textwidth]{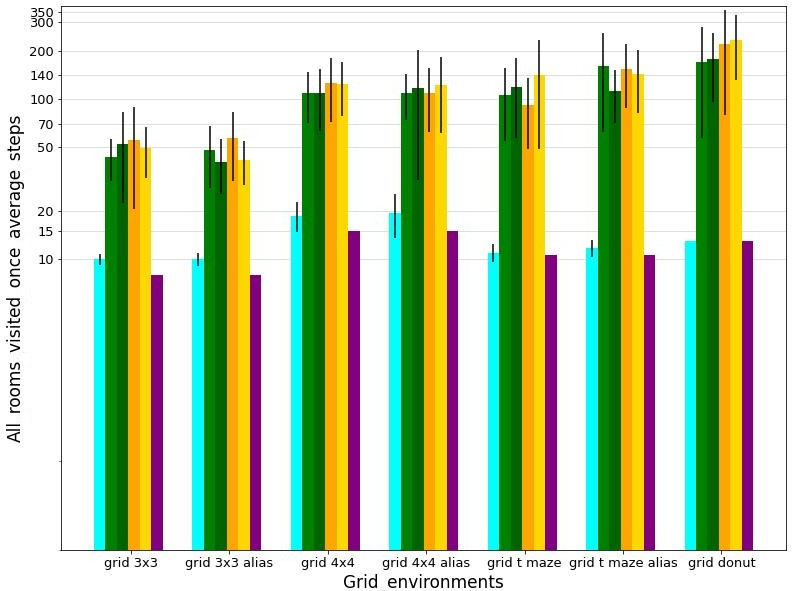}}

   \caption{The average steps are depicted on a logarithmic scale. Remarkably, our agent achieves all tasks in significantly fewer steps compared to the CSCG model. The oracle sets the benchmark, representing the minimum steps necessary to visit all rooms once. Additionally, an aliased room signifies the recurrence of identical observations across various locations, posing a challenge as it could mislead the agent regarding its current position.} 
\end{figure*}

To make the two models more similar for a fair comparison, we include our model's current state estimation mechanism in the CSCG approach~\cite{Toon_cognitive_maps}. Moreover, we decided to see how CSCG would behave if we included the position as an observation. This effectively removes aliasing and is believed equivalent to the proprioception of our model when starting without prior, we call that specific case "CSCG pose ob".
We compared the performance of our model (receiving only observation as input) to the CSCG receiving only visual observations or visual observation-position pairs with or without random exploration policies. The CSCG internal path estimator is based on the Viterbi method~\cite{viterbi,cscg_pres} and is updated every 5 steps with the sequence of pairs going from the first observation to the current time-step. 

Our environments are composed of several rooms connected in diverse ways (fully connected 3 by 3 and 4 by 4 rooms environments, T-shaped, donuts-shaped mazes with and without aliased floor colours). All models receive the room floor colour as observation. An example of a 3 by 3 rooms environment and extracted observations per room are presented in Figure~\ref{img:from_env_to_topo} first and second panel. The agents can move in the four cardinal directions or choose to stay at the present location. 

In our exploration runs, across all environments, agents are initially placed at random starting positions and tasked with learning the environment's topology. Results represent the mean over a minimum of ten successful runs in each environment. Notably, our agent always achieves successful exploration, while CSCG occasionally fails due to insufficient steps allotted to learn the topology. The Oracle model, analogous to an A-star path planning, demonstrates the ideal scenario where the agent seizes the full topology of the environment by visiting each position only once. In the case of the T-maze, results are averaged across all runs considering the starting positions of the models.
\begin{figure}[!htb]
    \centering
    \includegraphics[width=9cm]{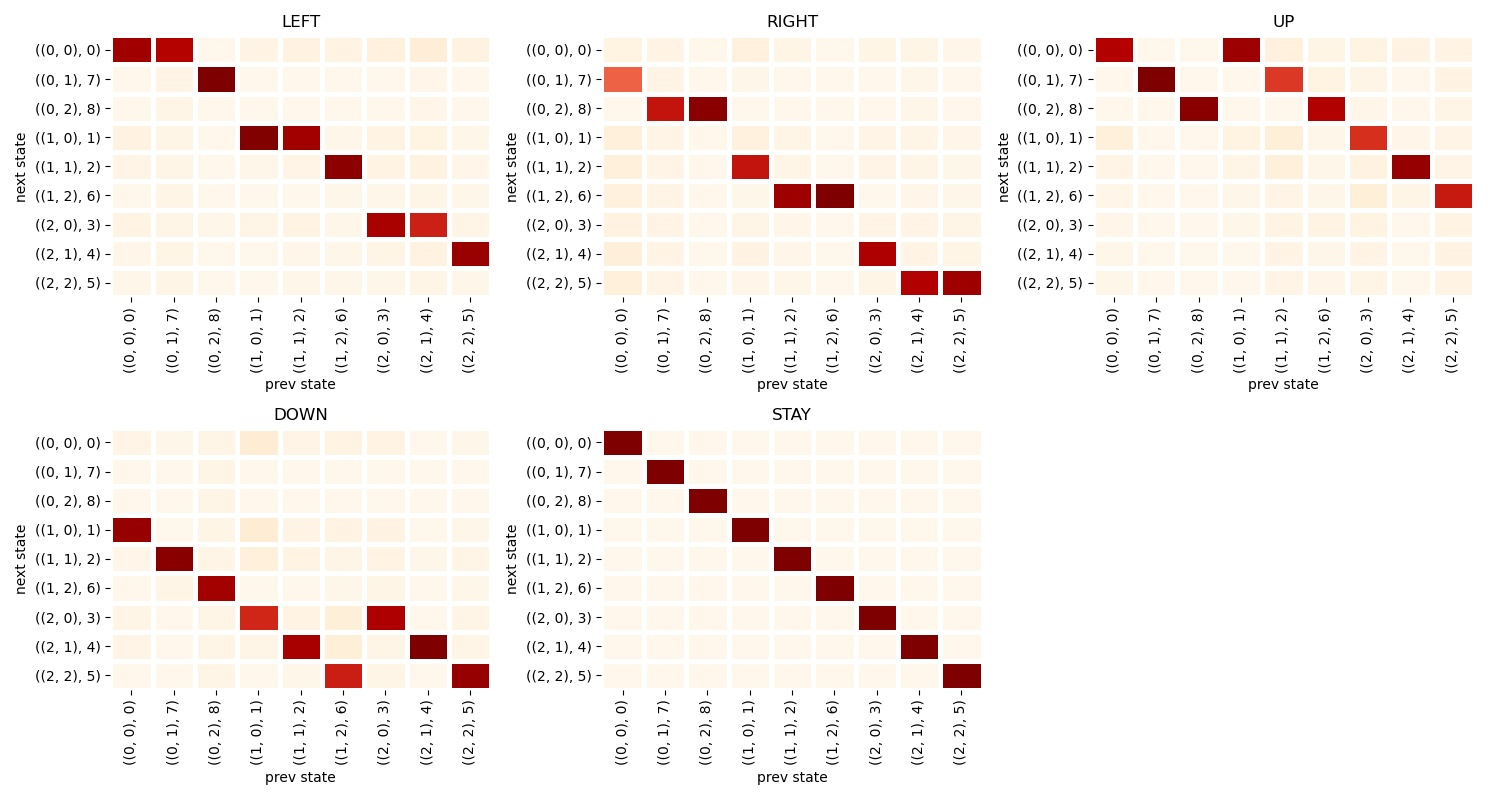}
    \caption{Example of a successful transition representation between positions in a 3x3 grid map. Each state in the plot is paired with its corresponding ground-truth pose for clarity (pose, state). The intensity of colour in the figure indicates the level of certainty the agent has about the transition.}
    \label{img:Transition_plot}
\end{figure}

Our exploration results can be seen in Figure~\ref{img:average_steps_models_per_env}. Exploration is deemed complete when the internal belief over the transition between observations aligns with the ground-truth transition matrix with a minimum certainty of 60\% overall correct transitions, the threshold was set arbitrarily based on the resulting successful transition representation as the one that can be seen in a 3 by 3 observations map depicted in figure~\ref{img:Transition_plot}. The figure shows how well-defined are possible transitions compared to impossible transitions (due to walls).  We also see that giving unique observations (visual observation-position pair) information reduces the CSCG training time of about 100 steps in aliased squared environments and 200 in T-shaped mazes, most probably due to its structure, the agent stays stuck in an aisle. However using random policy or the Viterbi algorithm for navigation does not improve exploration, because the agent can not extrapolate on unseen observations, thus finally leading to almost random action selection. This demonstrates the benefit of map extension over un-visited areas.

If we compare the number of steps required to learn the structure of the environment and the number of steps the agent takes to visit all unknown positions, we can deduce a few things.
Firstly, not having an imagination over possible trajectories disadvantages the CSCG, as it repeatedly visits known rooms, randomly or not, instead of being attracted to novelty as ours is. Ours has prior over non-visited states, rendering unknown rooms highly uncertain, and thus attractive to diminish the agent's internal model's parameters uncertainty.  
Secondly, we see our agent exploring all rooms with steps closely matching the oracle, this implies that it could have the potential to learn transitions faster, in a one-shot learning if we were to increase confidence in imagined beliefs.
However, this could also consolidate misbeliefs about transitions, in those experiments we let the agent confirm its priors instead of over-trusting them by setting the learning rate of the model low on predicted transitions. The given exploration seems to follow biological evidence on mouse behaviour in a maze~\cite{mice_in_labyrith}. 

\begin{figure}[!htb]
\centering
\subfloat[\label{img:agent_path_t_maze}]{\resizebox*{0.32\textwidth}{!}{\includegraphics{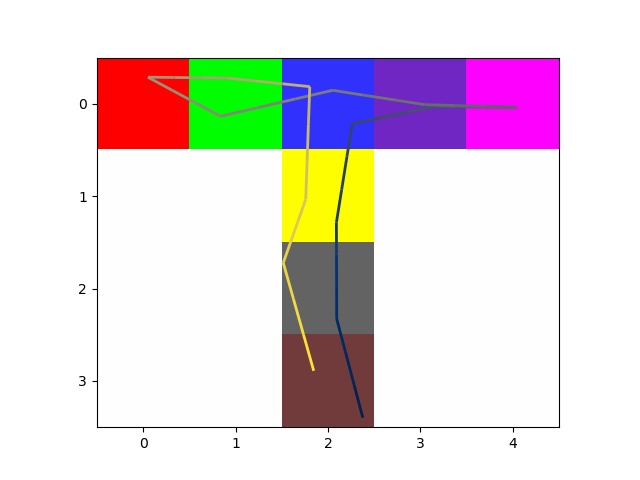}}}
\hspace{2pt}
\subfloat[\label{img:imagined_EFE}]{%
\resizebox*{0.63\textwidth}{!}{\includegraphics{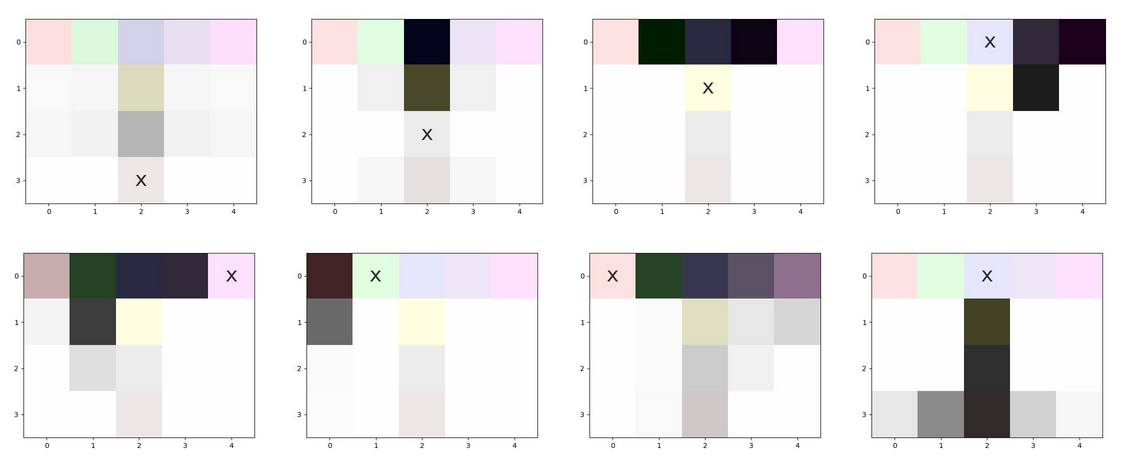}}}
   \caption{Exploration of a T-maze starting at the base of the T. a) depicts the full path as a line transitioning from black to white. b) showcases, from top to bottom columns one to two, the agent -represented as an X- imagined optimal policies. Darker colours indicate higher expected free energy.}
   \vspace{-2mm}
   \label{img:path_and_efe}
\end{figure}

We give a qualitative example of the agent behaviour in the T-maze Figure~\ref{img:path_and_efe}. Figure~\ref{img:agent_path_t_maze} shows the full path taken by a line varying from black to yellow, the agent starts at the bottom of the T-maze. Figure~\ref{img:imagined_EFE} shows the imagined trajectories of the agent (represented by X in the figure) at various steps to read from left to right and top to bottom. Imagined trajectories are associated with their expected free energy, the darker, the more desirable the path to the agent. The agent is purely driven by information gain in those experiments. Our model has low interest in paths leading into the current room walls and is highly attracted to unexplored areas.  Upon reaching the end of the right aisle (1st image, 2$^{nd}$ row of Fig~\ref{img:imagined_EFE}), the unexplored aisle is notably more attractive than the previously visited one, highlighting the agent's preference for uncertain observations over confirming existing beliefs. While returning to the starting point, the agent shows interest in paths going through walls. This is because these transitions become more intriguing as the agent has gained a better understanding of the environment's connectivity. A consolidation of its belief can be realised through a new observation of the walls.  Those observations confirm that the agent exhibits a coherent and effective exploration behaviour akin to how we would explore an environment, first discovering all areas before delving into specific details.

\FloatBarrier

\section{Discussion}

This study proposes a novel high-level abstraction model informed by biologically plausible principles mimicking key points of animal navigation strategies~\cite{Human_rodent_spatial_rep,humans-hierarchic-plan-subway}. By integrating a dynamic cognitive graph with internal positioning and an Active Inference Framework, our model successfully explores the environment and learns its structure in a few steps, as expected from animals~\cite{few_one_shot_learning,mice_in_labyrith}, facilitating adaptive learning and efficient exploration. Moreover, allowing the internal map to grow with expected beliefs not only creates a map adapted to any environment dimension, shape or observations but also enhances exploration by creating highly uncertain states where the whereabouts are predictable but the corresponding observations aren't. Comparative experiments with the Clone-Structured Graph (CSCG) model~\cite{cscg_pres} underscore the effectiveness of our approach in learning environment structures with minimal data and without prior knowledge of specific observation dimensions. This is mainly due to our agent's capacity to imagine actions' consequences and integrate them into its beliefs. Moving forward, it would be interesting to increase the prediction range of new states to integrate into the model and determine the impact on navigation. Moreover, studying the impact of a perfect memory on future policies and exploration efficiency, as well as seeing how the agent fares when trying to reach a defined objective it has prior upon in a familiar or novel environment would enhance the research. Finally deploying this model in real-world scenarios such as StreetLearn~\cite{street_learn}, based on Google map observations, would approach further this mechanism to animal behaviour and provide more conclusive evidence.

\section*{Acknowledgement}
This research received funding from the Flemish Government under the “Onder-
zoeksprogramma Artificiële Intelligentie (AI) Vlaanderen” programme.

\bibliographystyle{splncs04}
\bibliography{main.bib}

\begin{thebibliography}{10}
\providecommand{\url}[1]{\texttt{#1}}
\providecommand{\urlprefix}{URL }
\providecommand{\doi}[1]{https://doi.org/#1}

\bibitem{humans-hierarchic-plan-subway}
Balaguer, J., Spiers, H., Hassabis, D., Summerfield, C.: Neural mechanisms of hierarchical planning in a virtual subway network. Neuron  \textbf{90},  893--903 (05 2016). \doi{10.1016/j.neuron.2016.03.037}

\bibitem{grid_cell_nav}
Bush, D., Barry, C., Manson, D., Burgess, N.: Using grid cells for navigation. Neuron  \textbf{87},  507 -- 520 (2015), \url{https://api.semanticscholar.org/CorpusID:7275119}

\bibitem{gym_minigrid}
Chevalier-Boisvert, M., Willems, L., Pal, S.: Minimalistic gridworld environment for openai gym. \url{https://github.com/maximecb/gym-minigrid} (2018)

\bibitem{cscg_transformers}
Dedieu, A., Lehrach, W., Zhou, G., George, D., Lázaro-Gredilla, M.: Learning cognitive maps from transformer representations for efficient planning in partially observed environments (2024)

\bibitem{grid_place_cells_clustered_envs}
Edvardsen, V., Bicanski, A., Burgess, N.: Navigating with grid and place cells in cluttered environments. Hippocampus  \textbf{30} (08 2019). \doi{10.1002/hipo.23147}

\bibitem{hippo_cognitive_map}
Eichenbaum, H.: The hippocampus as a cognitive map … of social space. Neuron  \textbf{87}(1),  9--11 (2015). \doi{https://doi.org/10.1016/j.neuron.2015.06.013}, \url{https://www.sciencedirect.com/science/article/pii/S0896627315005267}

\bibitem{humans-cognitive-map}
Epstein, R., Patai, E.Z., Julian, J., Spiers, H.: The cognitive map in humans: Spatial navigation and beyond. Nature Neuroscience  \textbf{20},  1504--1513 (10 2017). \doi{10.1038/nn.4656}

\bibitem{AIF_learning}
Friston, K., FitzGerald, T., Rigoli, F., Schwartenbeck, P., Doherty, J.O., Pezzulo, G.: Active inference and learning. Neuroscience \& Biobehavioral Reviews  \textbf{68},  862--879 (2016). \doi{https://doi.org/10.1016/j.neubiorev.2016.06.022}, \url{https://www.sciencedirect.com/science/article/pii/S0149763416301336}

\bibitem{bayesian_model_reduc}
Friston, K., Parr, T., Zeidman, P.: Bayesian model reduction (2019)

\bibitem{cscg_pres}
George, D., Rikhye, R., Gothoskar, N., Guntupalli, J.S., Dedieu, A., Lázaro-Gredilla, M.: Clone-structured graph representations enable flexible learning and vicarious evaluation of cognitive maps. Nature Communications  \textbf{12} (04 2021). \doi{10.1038/s41467-021-22559-5}

\bibitem{human_exploration}
Gershman, S.: Deconstructing the human algorithms for exploration. Cognition  \textbf{173},  34--42 (12 2017). \doi{10.1016/j.cognition.2017.12.014}

\bibitem{cscg_abstraction}
Guntupalli, J.S., Raju, R., Kushagra, S., Wendelken, C., Sawyer, D., Deshpande, I., Zhou, G., Lázaro-Gredilla, M., George, D.: Graph schemas as abstractions for transfer learning, inference, and planning (02 2023). \doi{10.48550/arXiv.2302.07350}

\bibitem{aif_scene_construction}
Heins, R.C., Mirza, M.B., Parr, T., Friston, K., Kagan, I., Pooresmaeili, A.: Deep active inference and scene construction. Frontiers in Artificial Intelligence  \textbf{3} (2020). \doi{10.3389/frai.2020.509354}, \url{https://www.frontiersin.org/articles/10.3389/frai.2020.509354}

\bibitem{speed_cells}
Hinman, J., Brandon, M., Climer, J., Chapman, W., Hasselmo, M.: Multiple running speed signals in medial entorhinal cortex. Neuron  \textbf{91} (07 2016). \doi{10.1016/j.neuron.2016.06.027}

\bibitem{heading_cells_humans}
Jacobs, J., Kahana, M., Ekstrom, A., Mollison, M., Fried, I.: A sense of direction in human entorhinal cortex. Proceedings of the National Academy of Sciences of the United States of America  \textbf{107},  6487--92 (03 2010). \doi{10.1073/pnas.0911213107}

\bibitem{viterbi}
Jelinek, F.: Continuous speech recognition by statistical methods. Proceedings of the IEEE  \textbf{64}(4),  532--556 (1976). \doi{10.1109/PROC.1976.10159}

\bibitem{nav_aif}
Kaplan, R., Friston, K.: Planning and navigation as active inference  (12 2017). \doi{10.1101/230599}

\bibitem{SLAM_alias}
Lajoie, P., Hu, S., Beltrame, G., Carlone, L.: Modeling perceptual aliasing in {SLAM} via discrete-continuous graphical models. CoRR  \textbf{abs/1810.11692} (2018), \url{http://arxiv.org/abs/1810.11692}

\bibitem{Toon_cognitive_maps}
de~Maele, T.V., Dhoedt, B., Verbelen, T., Pezzulo, G.: Bridging cognitive maps: a hierarchical active inference model of spatial alternation tasks and the hippocampal-prefrontal circuit (2023)

\bibitem{street_learn}
Mirowski, P., Banki{-}Horvath, A., Anderson, K., Teplyashin, D., Hermann, K.M., Malinowski, M., Grimes, M.K., Simonyan, K., Kavukcuoglu, K., Zisserman, A., Hadsell, R.: The streetlearn environment and dataset. CoRR  \textbf{abs/1903.01292} (2019), \url{http://arxiv.org/abs/1903.01292}

\bibitem{weird_HAIF}
Neacsu, V., Mirza, M.B., Adams, R.A., Friston, K.J.: Structure learning enhances concept formation in synthetic active inference agents. PLOS ONE  \textbf{17}(11),  1--34 (11 2022). \doi{10.1371/journal.pone.0277199}, \url{https://doi.org/10.1371/journal.pone.0277199}

\bibitem{AIF_book}
Parr, T., Pezzulo, G., Friston, K.: Active Inference: The Free Energy Principle in Mind, Brain, and Behavior (03 2022). \doi{10.7551/mitpress/12441.001.0001}

\bibitem{map-graph-cognition}
Peer, M., Brunec, I.K., Newcombe, N.S., Epstein, R.A.: Structuring knowledge with cognitive maps and cognitive graphs. Trends in Cognitive Sciences  \textbf{25}(1),  37--54 (2021). \doi{https://doi.org/10.1016/j.tics.2020.10.004}, \url{https://www.sciencedirect.com/science/article/pii/S1364661320302503}

\bibitem{cscg_structuring_knowledge}
Peer, M., Brunec, I.K., Newcombe, N.S., Epstein, R.A.: Structuring knowledge with cognitive maps and cognitive graphs. Trends in Cognitive Sciences  \textbf{25}(1),  37--54 (2021). \doi{https://doi.org/10.1016/j.tics.2020.10.004}, \url{https://www.sciencedirect.com/science/article/pii/S1364661320302503}

\bibitem{cscg_space_latent_seq}
Raju, R.V., Guntupalli, J.S., Zhou, G., Lázaro-Gredilla, M., George, D.: Space is a latent sequence: Structured sequence learning as a unified theory of representation in the hippocampus (2022)

\bibitem{mice_in_labyrith}
Rosenberg, M., Zhang, T., Perona, P., Meister, M.: Mice in a labyrinth show rapid learning, sudden insight, and efficient exploration. eLife  \textbf{10},  e66175 (jul 2021). \doi{10.7554/eLife.66175}, \url{https://doi.org/10.7554/eLife.66175}

\bibitem{curiosity_exploitative}
Schwartenbeck, P., Passecker, J., Hauser, T.U., FitzGerald, T.H., Kronbichler, M., Friston, K.J.: Computational mechanisms of curiosity and goal-directed exploration. eLife  \textbf{8},  e41703 (may 2019). \doi{10.7554/eLife.41703}, \url{https://doi.org/10.7554/eLife.41703}

\bibitem{aif_step_by_step}
Smith, R., Friston, K.J., Whyte, C.J.: A step-by-step tutorial on active inference and its application to empirical data. Journal of Mathematical Psychology  \textbf{107},  102632 (2022). \doi{https://doi.org/10.1016/j.jmp.2021.102632}, \url{https://www.sciencedirect.com/science/article/pii/S0022249621000973}

\bibitem{AIF_modeling_struct}
Smith, R., Schwartenbeck, P., Parr, T., Friston, K.J.: An active inference approach to modeling structure learning: Concept learning as an example case. Frontiers in Computational Neuroscience  \textbf{14} (2020). \doi{10.3389/fncom.2020.00041}, \url{https://www.frontiersin.org/articles/10.3389/fncom.2020.00041}

\bibitem{border_cells}
Solstad, T., Boccara, C.N., Kropff, E., Moser, M.B., Moser, E.I.: Representation of geometric borders in the entorhinal cortex. Science  \textbf{322}(5909),  1865--1868 (2008). \doi{10.1126/science.1166466}, \url{https://www.science.org/doi/abs/10.1126/science.1166466}

\bibitem{ours_2024}
de~Tinguy, D., Van~de Maele, T., Verbelen, T., Dhoedt, B.: Spatial and temporal hierarchy for autonomous navigation using active inference in minigrid environment. Entropy  \textbf{26}(1), ~83 (Jan 2024). \doi{10.3390/e26010083}, \url{http://dx.doi.org/10.3390/e26010083}

\bibitem{few_one_shot_learning}
Tyukin, I.Y., Gorban, A.N., Alkhudaydi, M.H., Zhou, Q.: Demystification of few-shot and one-shot learning. CoRR  \textbf{abs/2104.12174} (2021), \url{https://arxiv.org/abs/2104.12174}

\bibitem{Human_rodent_spatial_rep}
Zhao, M.: Human spatial representation: What we cannot learn from the studies of rodent navigation. Journal of Neurophysiology  \textbf{120} (08 2018). \doi{10.1152/jn.00781.2017}

\end{thebibliography}

\end{document}